\documentclass[11pt,a4paper]{article}
\usepackage[hyperref]{acl2021}
\usepackage{times}
\usepackage{latexsym}

\usepackage{microtype}
\usepackage{amssymb}
\usepackage{amsmath}
\aclfinalcopy 
\def\aclpaperid{2266} 

\usepackage{amsmath}
\usepackage{enumitem}
\usepackage[T2A,T1]{fontenc}
\usepackage{color}
\usepackage{multirow}
\usepackage{graphicx}
\usepackage{enumerate}
\usepackage{booktabs}
\usepackage{subcaption}

\usepackage{algorithm}
\usepackage{algpseudocode}
\usepackage{algorithmicx,eqparbox,array}

\newcommand{\name}{GOT}

\title{Energy-based Unknown Intent Detection with Data Manipulation}

 \author{{\bf Yawen Ouyang},
  {\bf Jiasheng Ye},
  {\bf Yu Chen},
  {\bf Xinyu Dai}\textsuperscript{\thanks{$^*$ Corresponding author.}}, \\
  {\bf Shujian Huang} \and 
  {\bf Jiajun Chen} \\
  National Key Laboratory for Novel Software Technology, Nanjing University, China \\
  \texttt{\{ouyangyw,yejiasheng,cheny98\}@smail.nju.edu.cn} \\
  \texttt{\{daixinyu,huangsj,chenjj\}@nju.edu.cn} \\
}

\date{}

\begin{document}
\maketitle

\begin{abstract}


Unknown intent detection aims to identify the out-of-distribution (OOD) utterance whose intent has never appeared in the training set.
In this paper, we propose using energy scores for this task as the energy score is theoretically aligned with the density of the input and can be derived from any classifier. 
However, high-quality OOD utterances are required during the training stage in order to shape the energy gap between OOD and in-distribution (IND), and these utterances are difficult to collect in practice. To tackle this problem, we propose a data manipulation framework to \textbf{G}enerate high-quality \textbf{O}OD utterances with importance weigh\textbf{T}s (\textbf{\name}). Experimental results show that the energy-based detector fine-tuned by \name\, can achieve state-of-the-art results on two benchmark datasets.

\end{abstract}

\section{Introduction}
\label{intro}
Unknown intent detection is a realistic and challenging task for dialogue systems.
Detecting out-of-distribution (OOD) utterances is critical when employing dialogue systems in an open environment. It can help dialogue systems gain a better understanding of what they do not know, which prevents them from yielding unrelated responses and improves user experience.

A simple approach for this task relies on the softmax confidence score and achieves promising results ~\cite{hendrycks17baseline}. The softmax-based detector will classify the input as OOD if its softmax confidence score is smaller than the threshold. Nevertheless, further works demonstrate that using the softmax confidence score might be problematic as the score for OOD inputs can be arbitrarily high ~\cite{louizos2017multiplicative, NIPS2018}.

Another appealing approach is to use generative models to approximate the distribution of in-distribution (IND) training data and use the likelihood score to detect OOD inputs. However, ~\citet{ren2019likelihood} and \citet{gangal2019likelihood} find that likelihood scores derived from such models are problematic for this task as they can be confounded by background components in the inputs.

\begin{figure}[tbp]
    \centering
    \includegraphics[width=\columnwidth]{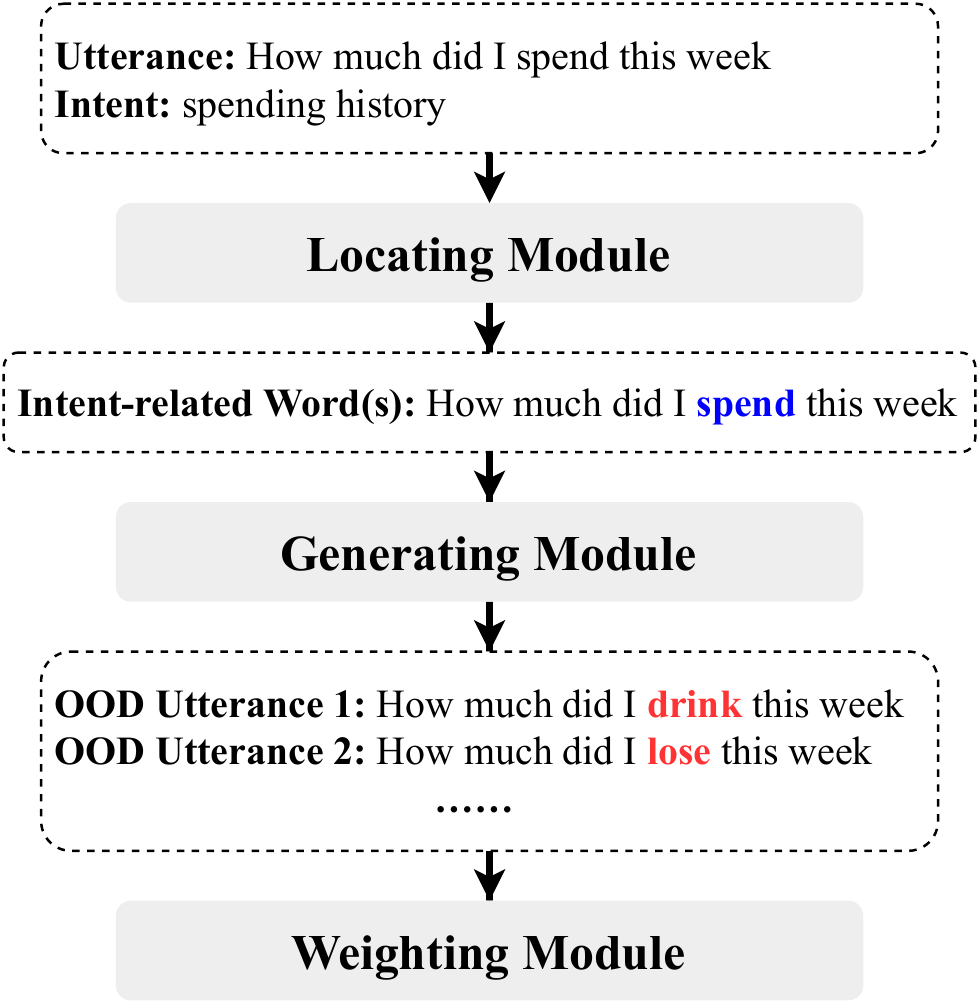}
    \caption{An overview of our framework \name. For the utterance ``How much did I spend this week'' from CLINC150 dataset ~\cite{larson-etal-2019-evaluation}. Our locating module locates the intent-related word ``spend''. And then our generating module generates words ``drink'', ``lose'' to replace it and obtains OOD utterances. Finally, our weighting module assigns a weight for each of OOD utterances.}
    \label{fig:flowchart}
\end{figure}

In this paper, we propose using \textit{energy scores} ~\cite{liu2020energy} for unknown intent detection. The benefit is that energy scores are theoretically well aligned with the density of the inputs, hence more suitable for OOD detection. Inputs with higher energy scores mean lower densities, which can be classified as OOD by the energy-based detector. Moreover, energy scores can be derived from any pre-trained classifier without re-training. Nevertheless, the energy gap between IND and OOD utterances might not always be optimal for differentiation. Thus we need auxiliary OOD utterances to explicitly shape the energy gap between IND and OOD utterances during the training stage ~\cite{liu2020energy}. 
This poses a new challenge in that the variety of possible OOD utterances is almost infinite. It is impossible to sample all of them to create the gap. 
~\citet{zheng2019out} demonstrate that OOD utterances akin to IND utterances, such as sharing the same phrases or patterns, are more effective, whereas these high-quality OOD utterances are difficult and expensive to collect in practice.

To tackle this problem, we propose a data manipulation framework \textbf{\name} \,to generate high-quality OOD utterances as well as importance weights.
\name \,generates OOD utterances by perturbing IND utterances locally, which allows the generated utterances to be closer to IND.
Specifically, \name \,contains three modules: (1) a locating module to locate intent-related words in IND utterances; (2) a generating module to generate OOD utterances by replacing intent-related words with desirable candidate words, evaluated in two aspects: whether the candidate word is suitable given the context, and whether the candidate word is irrelevant to IND; (3) a weighting module to reduce the weights of potential harmful generated utterances. Figure ~\ref{fig:flowchart} illustrates the overall process of \name. 
Experiments show that the generated weighted OOD utterances can further improve the performance of the energy-based detector in unknown intent detection. Our code and data will be available at: \url{https://github.com/yawenouyang/GOT}.

To summarize, the key contributions of the paper are as follows:
\begin{itemize}[leftmargin=*]
    \item We propose using energy scores for unknown intent detection. We conduct experiments on real-world datasets including CLINC150 and SNIPS to show that the energy score can achieve comparable performance as strong baselines.
    \item We put forward a new framework \name \,to generate high-quality OOD utterances and reweight them. We demonstrate that \name \, can further improve the performance of the energy score by explicitly shaping the energy gap and achieves state-of-the-art results.
    \item We show the generality of \name \, by applying generated weighted OOD utterances to fine-tune the softmax-based detector, and the fine-tuned softmax-based detector can also yield significant improvements.
\end{itemize}

\section{Related Work}
	
~\citealp{lane2006out}, ~\citealp{manevitz2007one} and ~\citealp{dai2007co} address OOD detection for the text-mining task.  Recently, this problem has attracted growing attention from researchers ~\cite{tur2014detecting, fei-liu-2016-breaking, fei2016learning, ryu2017neural, shu-etal-2017-doc}. ~\citet{hendrycks17baseline} present a simple baseline that utilizes the softmax confidence score to detect OOD inputs. ~\citet{shu-etal-2017-doc} create a binary classifier and calculate the confidence threshold for each class.  
Some distance-based methods ~\cite{oh2018out, lin-xu-2019-deep, yan-etal-2020-unknown} are also used to detect unknown intents as OOD utterances highly deviate from IND utterances in their local neighborhood. Simultaneously, with the advancement of deep generative models, learning such a model to approximate the distribution of training data is possible. However, ~\citet{ren2019likelihood} find that likelihood scores derived from these models can be confounded by background components, and propose a likelihood ratio method to alleviate this issue. ~\citet{gangal2019likelihood} reformulate and apply this method to unknown intent detection. 

Different from these methods, we introduce the energy score for this task. ~\citet{liu2020energy} prove that the energy score is theoretically aligned with the density of the input, and can be derived from any classifier without re-training, hence desirable for our task. We further propose a data manipulation framework to generate high-quality OOD utterances to shape the energy gap between IND and OOD utterances.

Note that there are some related works that also generate OOD samples to improve OOD detection performance. ~\citet{lee2017training} generate OOD samples with Generative Adversarial Network (GAN) ~\cite{NIPS2014}, and ~\citet{zheng2019out} explore this method for unknown intent detection. However, there are two major distinctions between our study and these works. First, they generate OOD utterances according to continuous latent variables, which cannot be easily interpreted. In contrast, our framework generates utterances by performing local replacements to IND utterances, which is more interpretable to human. Second, our framework additionally contains a weighting module to reform the generated utterances. Our work is also inspired by ~\citet{hengyi_acl20}, which proposes a framework to augment the IND data, while our framework aims to generate OOD data.
	
\section{Preliminary}

In this section, we formalize unknown intent detection task. Then we introduce the energy score, and its superiority and limitations for this task.
	
\subsection{Problem Formulation}
Given a training dataset $\mathcal{D}^{\rm{train}}_{\rm{in}} = \{(\mathbf{u}^{(i)}, y^{(i)})\}^N_{i=1}$ where $\mathbf{u}^{(i)}$ is an utterance and  $y^{(i)} \in Y_{\rm{in}} = \{y_1, y_2, ... , y_K\}$ is its intent label. In testing, given an utterance, unknown intent detection aims to detect whether its intent belongs to existing intents $Y_{\rm{in}}$. In general, unknown intent detection is an OOD detection task. The essence of all methods is to learn a score function that maps each utterance $\mathbf{u}$ to a single scalar that is distinguishable between IND and OOD utterances. 

\subsection{Energy-based OOD Detection}
An energy-based model ~\cite{lecun2006tutorial} builds an energy function $E(\mathbf{u})$ that maps an input $\mathbf{u}$  to a scalar called energy score (i.e., $E:\mathbb{R}^D\rightarrow \mathbb{R}$). Using the energy function, probability density $p(\mathbf{u})$ can be expressed as:
\begin{align}
p(\mathbf{u}) = \frac{\exp(-E(\mathbf{u})/T)}{Z}
\label{probablity_energy},
\end{align}
where $Z = \int_\mathbf{u}\exp(-E(\mathbf{u})/T)$ is the normalizing constant also known as the partition function and $T$ is the temperature parameter. Take the logarithm of both side of \eqref{probablity_energy}, we can get the equation:
\begin{align}
\log p(\mathbf{u}) = -E(\mathbf{u})/T - \log Z\label{log_p}.
\end{align}
Since $Z$ is constant for all input $\mathbf{u}$, 
we can ignore the last term $\log Z$  and find that the negative energy function $-E(\mathbf{u})$ is in fact linearly aligned with the log likelihood function, which is desirable for OOD detection ~\cite{liu2020energy}.\par
The energy-based model has a connection with a softmax-based classifier. For a classification problem with $K$ classes, a parametric function $f$ maps each input $\mathbf{u}$ to $K$ real-valued numbers (i.e., $f: \mathbb{R}^D \rightarrow \mathbb{R}^K$), known as logits. Logits are used to parameterize a categorical distribution using a softmax function:
\begin{align}
    p(y|\mathbf{u}) = \frac{\exp(f_y(\mathbf{u})/T)}{\sum_{y'} \exp(f_{y'}(\mathbf{u})/T)}\label{softmax},
\end{align}
where $f_y(\mathbf{u})$  indicates the $y^{\mathrm{th}}$  index of $f(\mathbf{u})$, i.e., the logit corresponding the  $y^{\mathrm{th}}$  class label. And these logits can be reused to define an energy function without changing function $f$ ~\cite{liu2020energy,grathwohl2019your}:
\begin{align}
    E(\mathbf{u}) = -T\cdot\log \sum_{y'} \exp{(f_{y'}(\mathbf{u})/T)}.
    \label{equ:energy}
\end{align}
According to the above, a classifier can be reinterpreted as an energy-based model. It also means the energy score can be derived from any classifier.

Due to its consistency with density and accessibility, 
we introduce the energy score for unknown intent detection, and utterances with higher energy scores can be viewed as OOD. Mathematically, the energy-based detector $G$ can be described as: 
\begin{align}
    G(\mathbf{u}; \delta, E) = 
    \begin{cases}
    \mbox{IND}&E(\mathbf{u})\le\delta,\\
    \mbox{OOD}&E(\mathbf{u})> \delta,
    \end{cases}
    \label{equ:detector}
\end{align}
where $\delta$ is the threshold.\par

Although the energy score can be easily computed from the classifier, the energy gap between IND and OOD samples might not always be optimal for differentiation. To solve this problem, ~\citet{liu2020energy} propose an energy-bounded learning objective to further widen the energy gap. Specifically, the training objective of the classifier combines the standard cross-entropy loss with a regularization loss:
\begin{align}
\mathcal{L}=\mathbb{E}_{(\mathbf{u},y)\sim \mathcal{D}^{\rm{train}}_{\rm{in}}} [-\log F_y(\mathbf{u})]+\lambda \cdot \mathcal{L}_{\rm{energy}},
\label{equ:all_loss}
\end{align}
where $F(\mathbf{u})$ is the softmax output, $\lambda$ is the auxiliary loss weight. The regularization loss is defined in terms of energy:
\begin{align}
    &\mathcal{L}_{\rm{energy}}=\mathbb{E}_{(\mathbf{u},y)\sim \mathcal{D}^{\rm{train}}_{\rm{in}}}(\max(0, E(\mathbf{u})-m_{\rm{in}}))^2\nonumber\\
    &+\mathbb{E}_{\hat{\mathbf{u}}\sim \mathcal{D}^{\rm{train}}_{\rm{out}}}(\max(0, m_{\rm{out}}-E(\hat{\mathbf{u}})))^2,
    \label{euq:regular_term}
\end{align}
which utilizes both labeled IND data $\mathcal{D}^{\rm{train}}_{\rm{in}}$ and auxiliary unlabeled OOD data $\mathcal{D}^{\rm{train}}_{\rm{out}}$. This term differentiates the energy scores between IND and OOD samples by using two squared hinge loss with the margin hyper-parameters $m_{\rm{in}}$ and $m_{\rm{out}}$.\par 

Ideally, one has to sample all types of OOD utterances to create the gap, which is impossible in practice. ~\citet{zheng2019out} demonstrate that OOD utterances akin to IND utterances could be more effective, but more difficult to collect. To address this problem, we propose a data manipulation framework, which can generate these high-quality OOD utterances and assign each generated utterance an importance weight to reduce the impact of potential bad generation.

\section{Approach}

In this section, we will introduce our data manipulation framework \name \, in detail. \name \, aims to generate high-quality OOD utterances by replacing intent-related words in IND utterances, and then assign a weight to each generated OOD utterance. Eventually, the weighted OOD utterances can be used to shape the energy gap.


\subsection{Locating Module}
\label{locating}
Since not all words in utterances are meaningful, such as stop words, when generating OOD utterances, replacing these words may not change the intent. It is more efficient and effective to replace those intent-related words. Hence, we design an intent-related score function $S$ to measure how a word $w$ related to an intent $y$:





\begin{align}
    \label{equ:locating}
    S(w, y) = \sum_{\mathbf{u} \in \mathcal{D}^{\rm{train}}_y} \sum_{w_j \in \mathbf{u}} &\mathbb{I}(w_j=w) [ \log p(w_j|\boldsymbol{w}_{<j},y) \nonumber \\
    &- \log p(w_j|\boldsymbol{w}_{<j})],
\end{align}
where $\mathcal{D}^{\rm{train}}_y$ is the subset of $\mathcal{D}^{\rm{train}}_{\rm{in}}$, which contains utterances with intent $y$, $\mathbb{I}$ is the indicator function, $w_j$ is the $j^{\rm{th}}$ word in $\mathbf{u}$, and $\boldsymbol{w}_{<j}=w_1,...,w_{j-1}$.


Given $w$ and $y$, the intent-related score function is the sum of the log-likelihood ratios for all $w$ in $\mathcal{D}^{\rm{train}}_y$. If $w$ is related to $y$, $w$ tends to occur more frequently in $\mathcal{D}^{\rm{train}}_y$ than other words. For each occurrence of $w$, i.e., $w_j$ equals $w$, $p(w|\boldsymbol{w}_{<j},y)$ should be higher than $p(w|\boldsymbol{w}_{<j})$ as the former is additionally conditioned on the related $y$, while the latter is not, hence resulting in a higher $S(w, y)$. In contrast, if $w$ is not related to $y$, $p(w|\boldsymbol{w}_{<j},y)$ is much less likely to be higher than $p(w|\boldsymbol{w}_{<j})$, or $w$ tends to have a lower frequency in $\mathcal{D}^{\rm{train}}_y$, hence $S(w, y)$ is likely to be small. Therefore, $S(w, y)$ can serve as a valid score function to measure how a word $w$ is related to an intent $y$. 

With the help of $S$, given an utterance to be replaced and its intent label, the locating module calculates the intent-related score for each word in this utterance, and a word with a higher score (i.e., larger than a given threshold) can be viewed as an intent-related word.

\begin{figure}[tbp]
    \centering
    \includegraphics[width=\columnwidth]{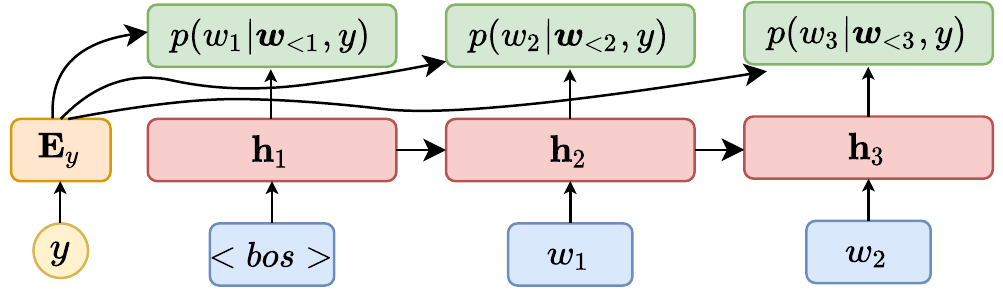}
    \caption{A class-conditional language model to estimate $p(w_j|\boldsymbol{w}_{<j},y)$.}
    \label{fig:cgm}
\end{figure}

\paragraph{Implementation: }We use two generative models to estimate $p(w_j|\boldsymbol{w}_{<j},y)$ and $p(w_j|\boldsymbol{w}_{<j})$ separately. Specifically, we train a class-conditional language model \cite{yogatama2017generative} with $\mathcal{D}^{\rm{train}}_{\rm{in}}$ to estimate $p(w_j|\boldsymbol{w}_{<j},y)$, shown in Figure ~\ref{fig:cgm}. To predict the word $w_j$, we can combine the hidden state $h_j$ with the intent embedding from a learnable label embedding matrix $\mathbf{E}_y$, then pass it through a fully connected (FC) layer and a softmax layer to estimate the word distribution. In the training process, the input is the utterance with its intent from $\mathcal{D}^{\rm{train}}_{\rm{in}}$, and the training objective is to maximize the conditional likelihood of utterances. 
To estimate $p(w_j|\boldsymbol{w}_{<j})$, we directly use pre-trained GPT-2 ~\cite{radford2019language} without tuning. Note that the whole training process only needs $\mathcal{D}^{\rm{train}}_{\rm{in}}$, and does not need auxiliary supervised data.

\subsection{Generating Module}

After detecting intent-related words in the utterance $\mathbf{u}$, for each of the intent-related words $w_t$, the generating module aims to generate the replacement words from the vocabulary set to replace $w_t$ and obtain OOD utterances. We design a candidate score function $Q$ to measure the desirability of the candidate word $c$:
\begin{equation}
    \begin{split}
        Q(c; \mathbf{u}, w_t) &= \log p(c|\boldsymbol{w}_{<t}, \boldsymbol{w}_{>t}) \\
        & - \log \sum_{y \in Y_{\rm{in}}} p(c|\boldsymbol{w}_{<t}, y)p(y).
    \end{split}
    \label{equ:candidate}
\end{equation}
The first term of the right hand side is the log-likelihood of $c$ conditioned on the context of $w_t$; the higher it is, the more suitable $c$ is given the context.
The second term of the right hand side is the negative log of the average likelihoods of $c$ conditioned on the IND label and previous context;
the higher it is, the less relevant $c$ is to IND utterances. Therefore, if $c$ has a higher candidate score, that means it fits the context well and has a low density under the IND utterance distribution, thus can be selected as the replacement word to replace $w_t$. The resulting generated OOD utterance is:
\begin{equation}
    \hat{\mathbf{u}} = \{\boldsymbol{w}_{<t}, c, \boldsymbol{w}_{>t}\}.
    \label{equ:generated}
\end{equation}

\paragraph{Implementation: }Similar with the locating module, we also do not need auxiliary supervised data to train the generating module. We use the same class-conditional language model mentioned in Section ~\ref{locating} to estimate $p(c|\boldsymbol{w}_{<t},y)$. $p(y)$ is the training set label ratios. To estimate $p(c|\boldsymbol{w}_{<t}, \boldsymbol{w}_{>t})$, we use pre-trained BERT ~\cite{devlin2018bert} without tuning.

\subsection{Weighting Module}

Since we cannot ensure the generation process is perfect, given a generated OOD utterance set $\mathcal{D}^{\rm{gen}}_{\rm{out}}=\{\hat{\mathbf{u}}^{(i)}\}^M_{i=1}$, there might be some unfavorable utterances that are useless or even harmful for tuning the classifier. To fit these utterances, the generalization ability of the classifier will decrease.
The weighting module aims to assign these utterances small weights. 

We first use Equation ~\ref{equ:all_loss} as the loss function to train a classifier by taking $\mathcal{D}^{\rm{gen}}_{\rm{out}}$ as $\mathcal{D}^{\rm{train}}_{\rm{out}}$. Then we calculate the influence value $\phi \in \mathbb{R}$ ~\cite{wang2020less} for each generated utterance $\hat{\mathbf{u}}$. The influence value approximates the influence of removing this utterance on the loss at validation samples. An utterance with positive $\phi$ implies that its removal will reduce the validation loss and strengthen the classifier's generalization ability, thus we should assign it a small weight. \footnote{Details about how to calculate the influence can be found in ~\cite{koh2017understanding, wang2020less}.} In particular, given $\phi$, we calculate weight $\alpha$ as follows:

\begin{equation}
    \alpha = \frac{1}{1+e^\frac{\gamma \phi}{\max_\phi - \min_\phi}},
    \label{equ:weight}
\end{equation}
where $\gamma \in \mathbb{R}^+$ is used to make the weight distribution flat or steep, $\max_\phi$ and $\min_\phi$ are the maximum and minimum influence value of utterances in $\mathcal{D}^{\rm{gen}}_{\rm{out}}$.

\paragraph{Implementation: } We still do not need auxiliary supervised data for this module. The validation loss is the cross-entropy loss on the validation set.

\begin{algorithm}[htbp]
  \small
  \caption{Data Manipulation Process}
  \label{alg: selection}
  \begin{algorithmic}[1]
    \Require
      Training set $\mathcal{D}^{\rm{train}}_{\rm{in}}$, intent-related score function $S$, candidate score function $Q$, intent-related word threshold $\epsilon$, candidate number K, weight term $\gamma$
    \Ensure
      Generated weighted OOD utterances set $\mathcal{D}^{\rm{gw}}_{\rm{out}}$
    
    \State $\mathcal{D}^{\rm{gen}}_{\rm{out}} = \{\}$
    \# generated OOD utterances without weights
    \For {$(\mathbf{u}, y) \in \mathcal{D}^{\rm{train}}_{\rm{in}}$}
        \For {$w_j \in \mathbf{u}$}
            \If {$S(w_j, y) > \epsilon$}
                \State $\mathcal{C} = \rm{top-}K_c \, Q(c;\mathbf{u}, w_j)$ 
                \For{$c \in \mathcal{C}$}
                    \State $\hat{\mathbf{u}} = \{\boldsymbol{w}_{<j}, c, \boldsymbol{w}_{>j}\}$
                    \State Add $\hat{\mathbf{u}}$ into $\mathcal{D}^{\rm{gen}}_{\rm{out}}$
                \EndFor
            \EndIf
        \EndFor
    \EndFor
    
    
    \State $\mathcal{D}^{\rm{gw}}_{\rm{out}} = \{\}$ \Comment{generated weighted OOD utterances}
    \For {$\hat{\mathbf{u}} \in \mathcal{D}^{\rm{gen}}_{\rm{out}}$}
        \State Calculate the weight $\alpha$ by Equation ~\ref{equ:weight}
        \State Add $(\hat{\mathbf{u}}, \alpha)$ into $\mathcal{D}^{\rm{gw}}_{\rm{out}}$
    \EndFor \\
    \Return{$\mathcal{D}^{\rm{gw}}_{\rm{out}}$}
  \end{algorithmic}
\end{algorithm}

\subsection{Overall Data Manipulation Process}

We summarize the process of \name \,
in Algorithm ~\ref{alg: selection}. Line 4 shows that $w_j$ can be viewed as an intent-related word for $y$ if $S(w_j, y)$ is greater than the intent-related word threshold $\epsilon$. Line 5 shows that we generate $K$ replacement words with the top-$K$ $Q(c;\mathbf{u}, w_t)$.

\subsection{Shape the energy gap with \name}

After obtaining weighted OOD utterances set $\mathcal{D}^{\rm{gw}}_{\rm{out}}$, we can explicitly shape the energy gap with them, resulting in IND utterances with smaller energy scores and OOD utterances with higher energy scores. Specifically, we redefine the regularization loss in Equation ~\ref{equ:all_loss} as follows and use it to re-train the classifier:
\begin{align}
    &\mathcal{L}_{\rm{energy}}=\mathbb{E}_{(\mathbf{u},y)\sim \mathcal{D}^{\rm{train}}_{\rm{in}}}(\max(0, E(\mathbf{u})-m_{\rm{in}}))^2\nonumber\\
    &+\mathbb{E}_{(\hat{\mathbf{u}}, \alpha) \sim \mathcal{D}^{\rm{gw}}_{\rm{out}}} \alpha (\max(0, m_{\rm{out}}-E(\hat{\mathbf{u}})))^2.
    \label{equ:new_en}
\end{align}

In the testing process, we can calculate the energy score for the utterance by Equation ~\ref{equ:energy}, and identify whether it is OOD by Equation ~\ref{equ:detector}.

\section{Experimental Setup}
	\label{Experimental Setup}
	
	\subsection{Datasets}
	To evaluate the effectiveness of the energy score and our proposed framework, we conducted experiments on two public datasets:
	
	\begin{itemize}[leftmargin=*]
		
		\item CLINC150\footnote{https://github.com/clinc/oos-eval} (\citealp{larson-etal-2019-evaluation}): this dataset covers 150 intent classes over ten domains. It supports some OOD utterances that do not fall into any of the system’s supported intents to avoid splitting unknown intents manually.
		\item SNIPS\footnote{https://github.com/snipsco/nlu-benchmark} (\citealp{coucke2018snips}): this dataset is a personal voice assistant dataset that contains seven intent classes. SNIPS does not explicitly include OOD utterances. We kept two classes \textit{SearchCreativeWork} and \textit{SearchScreeningEvent} as unknown intents.
		
	\end{itemize}
	
	Table ~\ref{tab:statistics} provides summary statistics about these two datasets. Note that the training set and validation set do not include OOD utterances.
	
	\begin{table}[t]
        \small
		\centering
			\begin{tabular}{lccc}
				\toprule
				Statistic & CLINC150 & SNIPS \\ 
				\midrule
				Train & 15000 & 9385 \\
				Validation & 3000 & 500 \\
				Test-IND & 4500 & 486 \\
				Test-OOD & 1000 & 214 \\
				Test-IND: Test-OOD & 4.5: 1 & 2.3: 1 \\
				Number of IND classes & 150 & 5 \\
				\bottomrule
			\end{tabular}
		\caption{Statistics of  CLINC150 and SNIPS datasets.} 
		\label{tab:statistics}
	\end{table}
	
	\begin{table*}[t]
		\centering
		\small
		\resizebox{\linewidth}{!}{
			\begin{tabular}{l|cccc|cccc}
				\hline
				\multirow{2}{4em}{Method} & \multicolumn{4}{c|}{CLINC150} &  \multicolumn{4}{c}{SNIPS}\\
				& \textbf{AUROC $\uparrow$} & \textbf{FPR95$\downarrow$} & \textbf{AUPR In$\uparrow$}  & \textbf{AUPR Out$\uparrow$} & \textbf{AUROC $\uparrow$} & \textbf{FPR95$\downarrow$} & \textbf{AUPR In$\uparrow$}  & \textbf{AUPR Out$\uparrow$} \\
				\hline
				MSP& 0.955 & 0.164 & 0.990 & 0.814 & 0.951 & 0.370 & 0.970 & 0.922 \\
				DOC& 0.943 & 0.221 & 0.985 & 0.790 & 0.938 & 0.493 & 0.956 & 0.910\\
				Mahalanobis & 0.969 & 0.118 & 0.993 & 0.871 & 0.979 & 0.088 & 0.989 & 0.964\\
				LMCL & 0.962 & 0.124 & 0.992 & 0.810 & 0.976 & 0.087 & 0.987 & 0.960\\
				SEG & 0.959 & 0.152 & 0.991 & 0.823 & 0.974 & 0.074 & 0.986 & 0.948  \\
				\hline
				Energy & 0.967 & 0.143 & 0.991 & 0.897 & 0.944 & 0.497 & 0.964 & 0.924 \\
				Energy + \name \, & \textbf{0.973}& \textbf{0.114} & \textbf{0.993}& \textbf{0.914} & \textbf{0.989}& \textbf{0.039}& \textbf{0.995}& \textbf{0.972}\\
				Energy + \name \,w/o weighting & 0.972 & 0.123 & 0.992 & 0.909 & 0.979 & 0.083 & 0.989 & 0.969 \\
				\hline
			\end{tabular}
		}
		\caption{AUROC, FPR95, AUPR In, AUPR Out on CLINC150, SNIPS datasets. Best results are in bold. All results are averaged across five seeds. }
		\label{main_result} 
	\end{table*}
	
	\subsection{Metrics}
	
	We used four common metrics for OOD detection to measure the performance. AUROC (\citealp{davis2006relationship}), AUPR In and AUPR Out (\citealp{manning1999foundations}) are threshold-independent performance evaluations and higher values are better. FPR95 is the false positive rate (FPR) when the true positive rate (TPR) is 95\%, and lower values are better.

	Considering the smaller proportion of OOD utterances in the test set on two datasets, AUPR Out is more informative here.
	
	\subsection{Baselines}
	
	\label{baselines}
	We introduce the following classifier-based methods as baselines: 

	\begin{itemize}[leftmargin=*]
    	\item MSP (\citealp{hendrycks17baseline}) trains a classifier with IND utterances and uses the softmax confidence score to detect OOD utterances.
		\item DOC (\citealp{shu-etal-2017-doc}) trains a binary classifier for each IND intent and uses maximum binary classifier output to detect OOD utterances.
		\item Mahalanobis (\citealp{NIPS2018}) trains a classifier with softmax loss and uses Mahalanobis distance of the input to the nearest class-conditional Gaussian distribution to detect OOD utterances.
		\item LMCL (\citealp{lin-xu-2019-deep}) uses LOF (\citealp{breunig2000lof}) in the utterance representation learned by a classifier. In training, they replace the softmax loss with LMCL (\citealp{wang2018additive}).
		\item SEG (\citealp{yan-etal-2020-unknown}) also uses LOF in the utterance representation. In training, they use semantic-enhanced large margin Gaussian mixture loss.
	\end{itemize}
	
	\subsection{Implementation Details}
	\label{details}
	For a fair comparison, all classifiers used in the above methods and ours are pre-trained BERT (\citealp{devlin2018bert}) with a multi-layer perceptron (MLP). 
	We select parameter values based on validation accuracy.
	For energy score, we follow \citet{liu2020energy} to set $T$ as 1, $\lambda$ as 0.1, $m_{\rm{in}}$ as -8 and $m_{\rm{out}}$ as -5. For influence value, we focus on changes on MLP parameters and use stochastic estimation \cite{koh2017understanding} with the scaling term 1000 and the damping term 0.003. For LMCL implementation, we set nearest neighbor number as 20, scaling factor s as 30 and cosine margin m as 0.35, which is recommended by \citet{lin-xu-2019-deep}. 
    For SEG, we follow \citet{yan-etal-2020-unknown} to set margin as 1 and trade-off parameter as 0.5.
	
	For our framework, we set candidate number $K$ as 2, weight term $\gamma$ as 20. In particular, for CLINC150, we set threshold $\epsilon$ as 150 and generate 100 weighted utterances for each intent. For SNIPS, we set threshold $\epsilon$ as 1500 and generate 1800 weighted utterances for each intent. The difference in settings between two datasets is due to the different sizes of per intent in the training set. 
	
	\section{Results and Analysis}
	In this part, we will show the results of different methods on two datasets and offer some further analysis.
	
	\subsection{Overall Results}
	\label{Overall Results}
	As shown in Table ~\ref{main_result}, we can observe that: 
	\begin{itemize}[leftmargin=*]
	\item The energy score can achieve comparable results on two datasets. Note that on SNIPS dataset, the advantages of the energy score are not as obvious. The reason is that SNIPS dataset is not as challenging as CLINC150 dataset, most methods can achieve good results, such as AUPR out is greater than 0.9.
    \item Energy + \name \, achieves better results on two datasets as compared to the raw energy score. It indicates that our generated weighted OOD utterances can effectively shape the energy gap, resulting in more distinguishable between IND and OOD utterances.
    \item We also report ablation study results. ``w/o weighting'' is the energy score tuned by OOD utterances without reweighting. We can see that there is a decrease in performance on both datasets, which shows the advantage of the weighting module (p-value < 0.005).
    \end{itemize}
    
	
    \begin{figure}[tbp]
    \centering
    \includegraphics[width=\columnwidth]{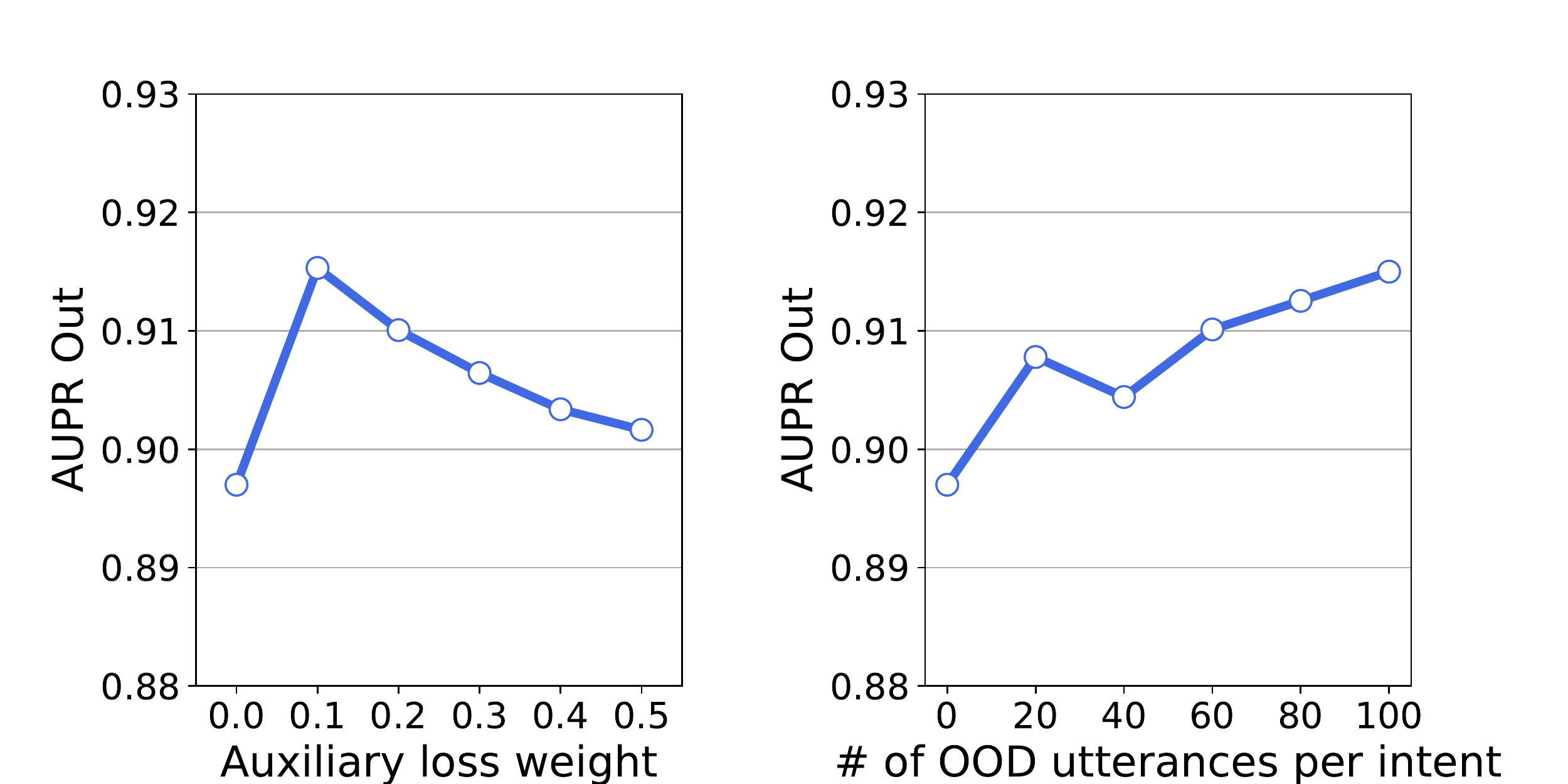}
    \caption{Effect of the auxiliary loss weight (left) and the number of generated weighted OOD utterances per intent (right).}
    \label{fig:eff}
    \end{figure}

    \subsection{Effect of Hyper-parameters}
    \label{Effect of Hyperparameters}
    During the training process, we find that the method performance is sensitive to two hyper-parameters: auxiliary loss weight $\lambda$ and the number of generated weighted OOD utterances per intent. We conduct two experiments to demonstrate their effects separately. We choose CLINC150 dataset as it is more challenging as mentioned before. \par
    \paragraph{Auxiliary Loss Weight:} We set the auxiliary loss weight $\lambda$ from $0$ to $0.5$ with an interval of $0.1$ to observe its impact.\par
    Results are shown in Figure ~\ref{fig:eff} (left). With the increase of auxiliary loss weight, the performance increases first and then decreases. $\lambda=0.1$ achieves the highest AUPR Out $0.914$ and outperforms $\lambda=0$ with an improvement of 1.7\% (AUPR Out). The results suggest that although shaping the energy gap can improve the performance, there exists a trade-off between optimizing the regularization loss and optimizing cross-entropy loss.\par
    \paragraph{Number of OOD Utterances:} we compare the performance of generated weighted utterance numbers for each intent by adjusting the number from $0$ to $100$ with an interval of $20$.\par
    Results are shown in Figure ~\ref{fig:eff} (right). As a whole, AUPR Out increases as more OOD utterances are incorporated into training. We can see that the performance is also improved even with a small generated number, which indicates the necessity of explicitly shaping the energy gap.

    \begin{table}[tbp]
		\centering
		\small
		\resizebox{\columnwidth}{!}{
			\begin{tabular}{l|cccc}
				\hline
				Method
				& \textbf{AUROC $\uparrow$} & \textbf{FPR95$\downarrow$} & \textbf{AUPR In$\uparrow$}  & \textbf{AUPR Out$\uparrow$} \\
				\hline
				Energy & 0.967 & 0.143 & 0.991 & 0.897 \\
				Energy + Wiki& 0.961 & 0.170 & 0.988 & 0.889 \\
				\hline
			\end{tabular}
		}
		\caption{Effect of using Wikipedia sentences to shape the energy gap.} 
		\label{tr_comp_result} 
	\end{table}
	
	\begin{table}[tbp]
		\centering
		\small
		\resizebox{\columnwidth}{!}{
			\begin{tabular}{l|cccc}
				\hline
				Method 
				& \textbf{AUROC $\uparrow$} & \textbf{FPR95$\downarrow$} & \textbf{AUPR In$\uparrow$}  & \textbf{AUPR Out$\uparrow$} \\
				\hline
				MSP& 0.955 & 0.164 & 0.990 & 0.814 \\
				MSP + \name \, & 0.972 & 0.118 & 0.993 & 0.903 \\
				\hline
			\end{tabular}
		}
		\caption{Effect of using \name \,to fine-tune the softmax-based detector.} 
		\label{msp_result} 
	\end{table}

    \begin{table*}[!tbp]
    \centering
    \small
    \resizebox{\linewidth}{!}{
     \begin{tabular}{llcc}
      \hline
    	\textbf{Intent} & \textbf{IND utterance and intent-related word [\textbf{w}]} &\textbf{Replacement word}&\textbf{Weight}\\
      \hline
      \multirow{2}*{Insurance}
      &i need to know more about my health [\textbf{plan}]&problems&0.50\\
      &what [\textbf{benefits}] are provided by my insurance & services &0.19\\
    \hline
    \multirow{2}*{Credit Limit Change}&can i get a higher limit on my american express [\textbf{card}]&ticket&0.46\\
    &can you [\textbf{increase}] how much i can spend on my visa&guess&0.54\\
    \hline
    \multirow{2}*{Reminder}&can you list each item on my [\textbf{reminder}] list&contacts&0.50\\
    &what's on the [\textbf{reminder}] list&agenda&0.78\\
     \hline
     \multirow{2}*{Redeem Rewards}&walk me through the process of cashing in on [\textbf{credit}] card points&those&0.22\\
     &i have credit card [\textbf{points}] but don't know how to use them&privileges&0.50 \\
     \hline
      \end{tabular}
    }
      \caption{Weighted OOD utterances generated by \name \, on CLINC150 dataset.}
    \label{case_study}
    \end{table*}
    
    \subsection{Compare with Wikipedia Sentences}
    
    An easy way to obtain OOD utterances is from the Wikipedia corpus.
    We investigate the effect of regarding Wikipedia sentences as OOD utterances to shape the energy gap on CLINC150 dataset. The Wikipedia sentences are from ~\citet{larson-etal-2019-evaluation} and the number is 14750.
    
    As shown in Table ~\ref{tr_comp_result}, we can observe that these sentences cannot improve the performance and even have a negative effect (We experimented with several hyper-parameters, this is the best result we could get). After observing these Wikipedia sentences, we find that they have little relevance to IND utterances. Therefore, simply using Wikipedia sentences is unrepresentative and ineffective for shaping the energy gap. 
    
    
    \subsection{\name \, for Softmax-based Detector}
    As mentioned in Section ~\ref{intro}, when using the softmax-based detector, OOD inputs may also receive a high softmax confidence score. 
    To tackle this problem, ~\citet{lee2017training} replace the cross entropy loss with the confidence loss. The confidence loss adds the Kullback-Leibler loss (KL loss) on the original cross entropy loss, which forces OOD inputs less confident by making their predictive distribution to be closer to uniform.
    
\begin{figure}[tbp]
    \centering
    \includegraphics[width=\columnwidth]{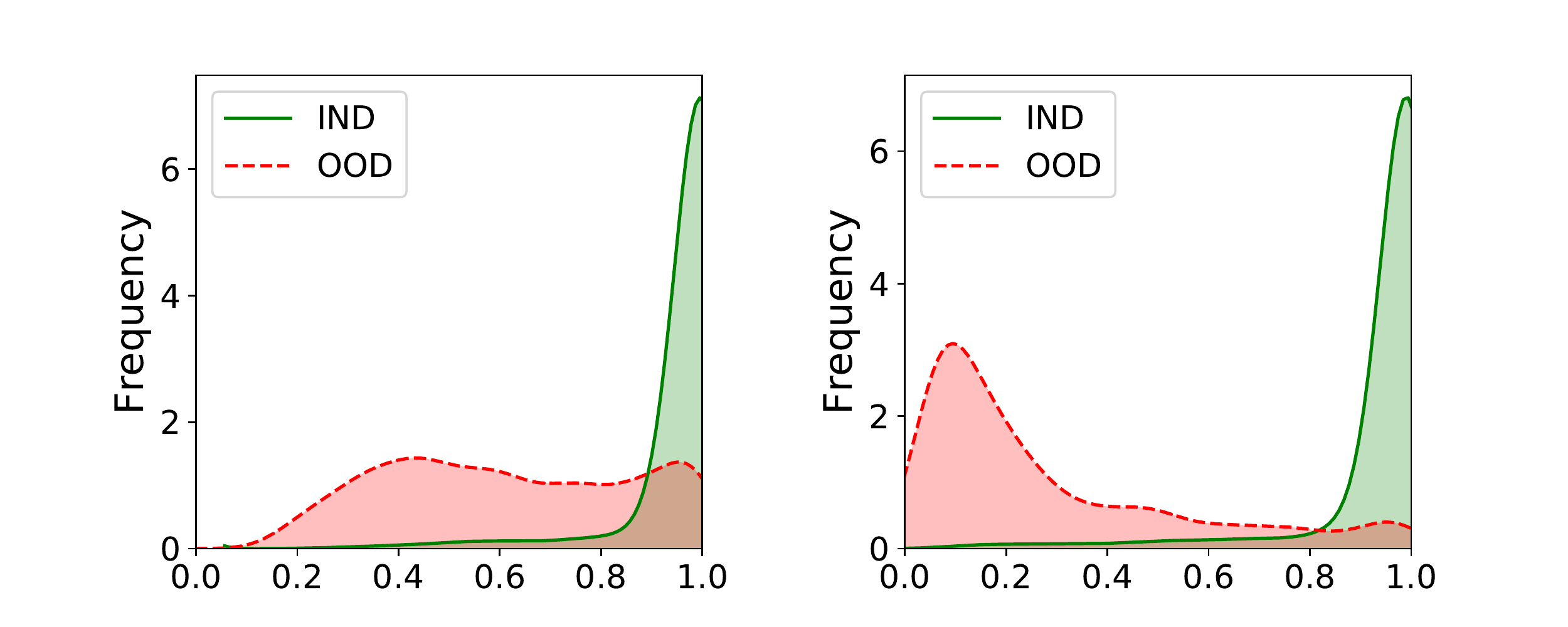}
    \caption{Histogram of the softmax confidence score from MSP (left) and MSP + \name \,(right).}
    \label{fig:density}
    \end{figure}

    To verify the generality of \name, we directly use the generated weighted OOD utterances to fine-tune the softmax-based detector with the confidence loss. The results are shown in Table ~\ref{msp_result}. Our MSP + \name \,has a significant improvement and outperforms MSP by 8.9\% (AUPR Out). Figure ~\ref{fig:density} provides an intuitive presentation. The softmax confidence scores of OOD from MSP form smooth distributions (see Figure ~\ref{fig:density} (left)). In contrast, the softmax confidence scores of OOD from MSP + \name \,concentrate on small values (see Figure ~\ref{fig:density} (right)). Overall the softmax confidence score is more distinguishable between IND and OOD after tuning by \name.
    
 \subsection{Case Study for \name}


We sample some intents and showcase generated weighted OOD utterances in Table \ref{case_study}. We can observe that intent-related words that located by our locating module are diverse, containing not only words appeared in the intent label. The replacement word fits the context well, and the intent of the generated utterance is exactly changed in most conditions. Admittedly, {\name} may have a bad generation, like replace ``benefits'' with ``services'' in the second utterance, which leads the generated utterance is still in-domain. Fortunately, the weighting module assigns these utterances a lower weight to reduce their potential harm.

\section{Conclusion and Future Work}

In this paper, we propose using energy scores for unknown intent detection and provide empirical evidence that the energy-based detector is comparable to strong baselines. To shape the energy gap, we propose a data manipulation framework \name \,to generate high-quality OOD utterances and assign their importance weights. We show that the energy-based detector tuned by \name \,can achieve state-of-the-art results. We further employ generated weighted utterances to fine-tune the softmax-based detector and also achieve improvements.




In the future, we will explore more operations, such as insertion, drop, etc., to enhance the diversity of generated utterances.

\section*{Acknowledgments}
We would like to thank the anonymous reviewers for their constructive comments. This work was supported by NSFC Projects (Nos. 61936012 and 61976114), the National Key R\&D Program of China (No. 2018YFB1005102).
	
\bibliographystyle{acl_natbib}
\bibliography{acl2021}

\end{document}